\documentclass[11pt]{article}

\usepackage[preprint]{acl}

\usepackage{times}
\usepackage{latexsym}

\usepackage{enumitem}
\usepackage{colortbl}
\usepackage[table]{xcolor}

\usepackage{subcaption}

\usepackage{dsfont}

\usepackage{amsmath}
\usepackage{amssymb}
\usepackage{algorithm}
\usepackage{algorithmic} 

\usepackage{booktabs} 

\usepackage{graphicx} 

\usepackage{multirow}

\usepackage{algorithm}
\usepackage{amsmath}
\usepackage{amssymb} 

\usepackage[T1]{fontenc}

\usepackage[utf8]{inputenc}

\usepackage{microtype}

\usepackage{inconsolata}

\usepackage{graphicx}

\title{CascadeDebate: Multi-Agent Deliberation for Cost-Aware LLM Cascades}

\author{Raeyoung Chang\thanks{Equal Contribution.}\\
  Sogang University \\
  \texttt{icanry@sogang.ac.kr} \And
  Dongwook Kwon\footnotemark[1] \\
  Kwangwoon University \\
  \texttt{dongwook.kwon@kw.ac.kr} \AND
  Jisoo Lee\footnotemark[1] \\
  Seoul National University \\
  \texttt{sally66890@snu.ac.kr} \And
  Nikhil Verma\thanks{Project Lead and corresponding author.} \\
  LG Electronics, Toronto AI Lab \\
  \texttt{nikhil.verma@lge.com}}

\begin{document}
\maketitle

\begin{abstract}
Cascaded LLM systems coordinate models of varying sizes with human experts to balance accuracy, cost, and abstention under uncertainty. 
However, single-model tiers at each stage falter on ambiguous queries, triggering premature escalations to costlier models or experts due to under-confidence and inefficient compute scaling.
CascadeDebate addresses this critical gap by inserting multi-agent deliberation directly at each tier's escalation boundary. 
Confidence-based routers activate lightweight agent ensembles only for uncertain cases, enabling consensus-driven resolution of ambiguities internally, without invoking higher-cost upgrades.
Our unified architecture alternates single-model inference with selective multi-agent deliberation across model scales, culminating in human experts as final fallback. 
This design scales test-time compute dynamically to query difficulty.
Across five benchmarks spanning science, medicine, and general knowledge, CascadeDebate outperforms strong single-model cascades and standalone multi-agent systems by up to 26.75\%.
An online threshold optimizer proves essential, boosting accuracy 20.98–52.33\% relative improvement over fixed policies and enabling elastic adaptation to real-world distributions.
\end{abstract}

\section{Introduction}

Large language models~(LLMs) have demonstrated remarkable proficiency across diverse benchmarks, spanning scientific question answering to medical diagnosis tasks~\citep{hendrycks2021measuring, pal2022medmcqa, singhal2023large}. 
Despite these advances, real-world deployment demands far more than raw predictive capability.
The systems must balance high accuracy against substantial inference costs, particularly as applications scale to production environments~\citep{chen2024frugalgpt}. 
High-capacity models deliver superior performance, but incur prohibitive computational overhead during training and inference~\citep{yue2024large}. 
Compact alternatives promise efficiency, yet sacrifice reliability, remaining prone to hallucinations and breakdowns in multi-step reasoning~\citep{wei2022chain, ji2023survey}.

\begin{figure}[t]
    \raggedright
    \includegraphics[width=0.45\textwidth]{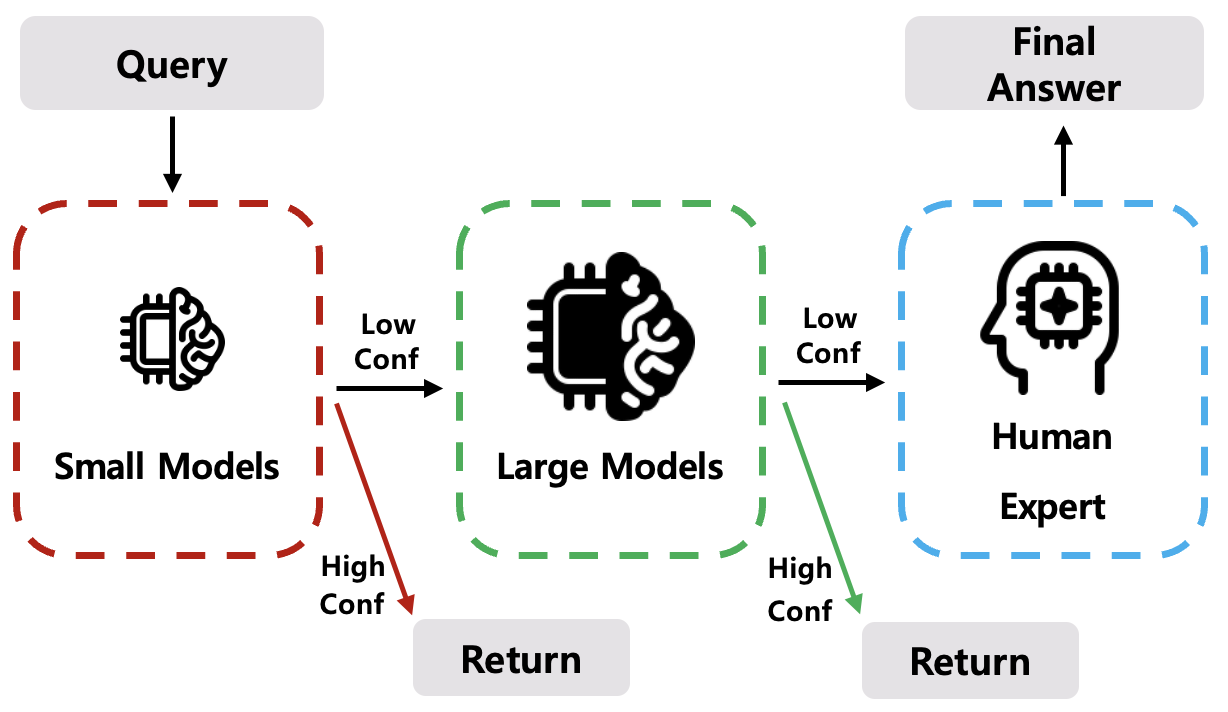}
    \caption{\textbf{Adaptive Cascade Framework Overview.} Current systems route queries through a hierarchy of solvers: from small models to large models, finally to human experts. High-confidence answers commit immediately at each stage; uncertain cases escalate to balance cost and accuracy.}
    \label{fig:overall_motivation}
\end{figure}

As shown in Figure~\ref{fig:overall_motivation}, current approaches mitigate this tension through cascaded pipelines and routing mechanisms~\citep{aggarwal2024automix, ong2025routellm}. 
These systems route queries from smaller models to larger ones based on confidence or acceptance criteria, deferring only when uncertainty signals potential failure. 
While effective for gross cost-accuracy trade-offs, such cascades rely on single-model decisions at each tier.
Ambiguous inputs thus trigger premature escalations, propagating overconfident errors without intra-tier correction, and wasting compute on unnecessary upgrades~\citep{fanconi2025cascaded}.
Multi-agent frameworks offer a compelling counterpoint.
They harness collective intelligence through debate, consensus, or role specialization to boost reasoning depth~\citep{du2024improving, li2024more, li2023camel}. 
Agents deliberate collaboratively, surfacing diverse perspectives that resolve ambiguities a lone model might mishandle. 
Yet these setups typically operate in isolation, as standalone enhancers rather than embedded components within cost-aware hierarchies~\citep{wang2025mixtureofagents}. 
Standalone multi-agent systems overlook selective activation: deliberation excels on hard cases, but proves redundant and costly for straightforward queries.

\begin{figure*}[t]
    \centering
    \includegraphics[width=0.85\textwidth]{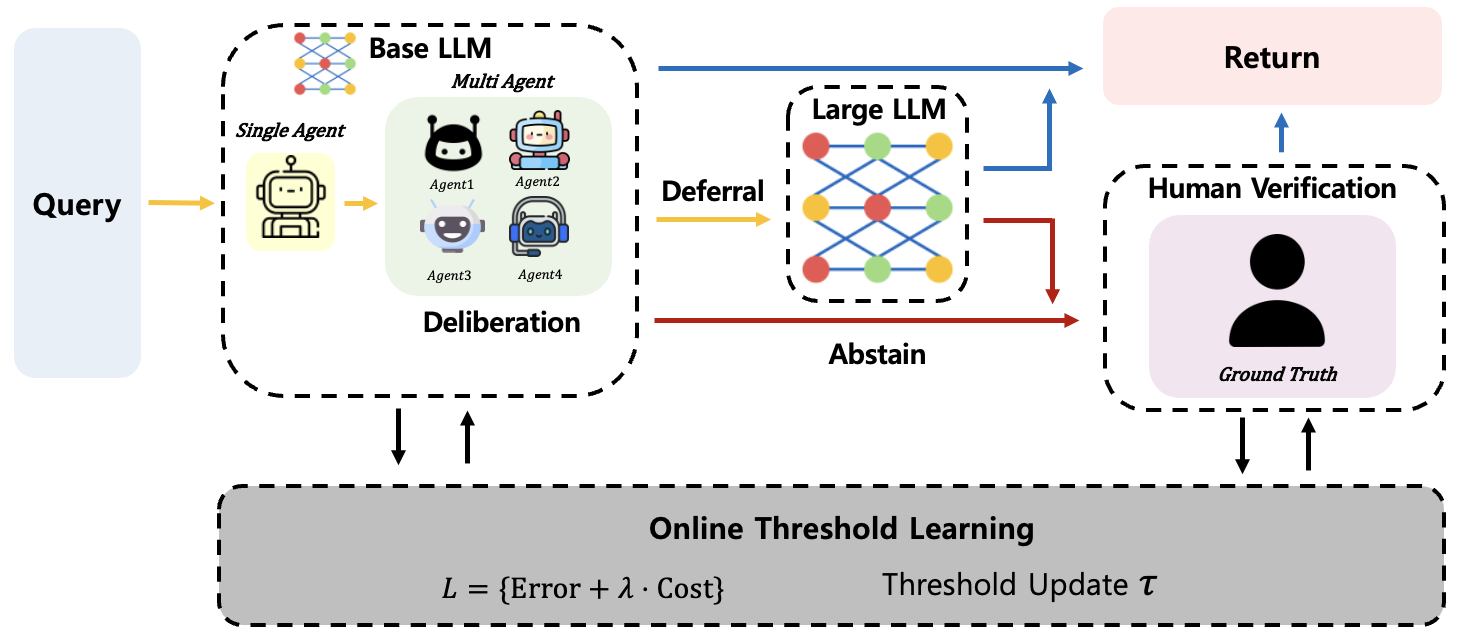}
    \caption{\textbf{CascadeDebate Architecture.} 
    The unified framework alternates single-model inference with multi-agent deliberation across base and large scales, culminating in human experts as final fallback. 
    Confidence-based routers activate lightweight agent ensembles only for marginal cases at each escalation boundary, resolving ambiguities via intra-tier consensus before deferring to costlier tiers. 
    The online threshold optimizer continuously refines deferral boundaries $\boldsymbol{\tau}$ from human feedback, enabling elastic adaptation to real-world query distributions while balancing accuracy against token and expert costs.
    }

    \label{fig:overall_arch}
\end{figure*}

This gap calls for a hybrid approach that embeds multi-agent deliberation within cascades, triggered only at escalation boundaries. 
Instead of routing uncertain queries from a single model directly to expensive larger tiers, we position agent consensus as an intermediate step. 
Such intra-tier scaling harnesses emergent cooperation. It self-corrects for ambiguities before any escalation occurs. 
This preserves efficiency while strengthening base models. 
Agents supply the diverse viewpoints missing from lone models, exactly where current cascades squander compute through premature deferrals. 
We present CascadeDebate, a unified architecture alternating single-model inference with multi-agent deliberation across scales~(shown in Figure~\ref{fig:overall_arch}). 
Human experts serve as final fallback. 
Confidence-based routers control progression. High-confidence single-model outputs commit right away.
Marginal cases activate lightweight agent ensembles for consensus refinement.
Only unresolved uncertainty advances to larger tiers or humans.

This design embodies test-time compute scaling tailored to query difficulty where deliberation depth emerges dynamically from need rather than fixed policy.
To ensure adaptability without manual tuning, we introduce an online optimizer that refines escalation thresholds from streaming human feedback. 
The mechanism learns task-specific boundaries online, minimizing token costs while maximizing accuracy, transforming the cascade into an elastic reasoner responsive to real-world query distributions.
Our contributions advance this vision:

\begin{itemize}
    \item A unified cascade architecture interposing multi-agent deliberation at each tier's escalation boundary, with human-in-the-loop as ultimate recourse.
    \item An online threshold learner balancing accuracy against computation, enabling continual refinement from human signals.
    \item Empirical validation showing that adaptive stage selection outperforms single-model cascades and standalone agents, resolving ambiguities internally to reduce expert load substantially.
\end{itemize}

This architecture not only bridges cascades and multi-agent paradigms but deploys practical human-AI collaboration at scale. 
Subsequent sections detail the architecture (Sec.~\ref{sec:methodology}), experiment setup~(Sec.~\ref{sec:setup}), and rigorous evaluation~(Sec.~\ref{sec:results}).

\section{Related Work}

\subsection{Cost-Aware Routing and LLM Cascades}
Adaptive inference systems optimize cost-accuracy trade-offs by routing queries across model tiers \citep{chen2024frugalgpt, aggarwal2024automix}.
Cascaded architectures prioritize lightweight models and escalate only when acceptance criteria are not met \citep{ong2025routellm, ding2024hybrid, yue2024large}.
These frameworks typically use single-model inference at each decision point, which makes routing sensitive to noisy confidence estimates and calibration error \citep{xiong2024can, fanconi2025cascaded}.
CascadeDebate addresses this limitation by adding selective intra-tier computation that targets marginal cases before inter-tier escalation.

\subsection{Multi-Agent Deliberation and Consensus}
Multi-agent systems improve reasoning robustness through debate, critique, and role specialization, followed by aggregation \citep{du2024improving, lee2025gemmas, li2023camel, li2024more}.
These mechanisms reduce single-model variance and expose failure modes that isolated inference may miss \citep{zhang2024chain, wang2025mixtureofagents}.
Prior work also reduces coordination overhead through sparse interaction structures or structured roles \citep{SunZW025_cortex, chen2025multi}.
However, deliberation is typically deployed as a standalone enhancement rather than as a selective operator integrated with cascade routing decisions.
CascadeDebate integrates deliberation at escalation boundaries to align additional computation with uncertainty.


\subsection{Online Threshold Learning}
LLM confidence estimates are poorly calibrated \citep{xiong2024can}, making static routing thresholds brittle under distribution shifts and heterogeneous domain difficulty,  motivating adaptive deferral policies \citep{fanconi2025cascaded}.
Unified formulations connect routing and cascading to support learning deferral behavior instead of manual threshold selection \citep{dekoninck2024unified, shen2025sater}.
CascadeDebate follows this direction by updating escalation thresholds online from streaming human feedback. 
The resulting policy adapts stage utilization to the observed workload while preserving explicit control over the cost-accuracy trade-off.

\section{Methodology}
\label{sec:methodology}

CascadeDebate extends cascaded decision systems by embedding multi-agent deliberation at escalation boundaries between model tiers. 
We denote the cascade as $C = \{S_1, S_2, \ldots, S_K\}$, where each stage $S_k$ processes input $x \in X$ to produce answer $\hat{y}_k \in Y$, confidence $\Phi_k(x) \in [0,1]$, and uncertainty $\Xi_k(x) \in [0,\infty)$, following terminology of~\cite{fanconi2025cascaded}.
At each stage, progression follows the deferral policy: accept $\hat{y}_k$ if $\Phi_k(x) > \tau^d_k$ (confidence exceeds deferral threshold); abstain to humans if $\Xi_k(x) > \tau^a_k$; otherwise defer to $S_{k+1}$.
%
%
For basic cascade we follow two-tier model architecture $M_\text{base} \to M_\text{large}$, but it can be generalized to arbitrary depth $K$.

\subsection{CascadeDebate architecture}
It alternates single-model inference $S_\text{single}$ with multi-agent deliberation $S_\text{multi}$ across model scales. 
Formally, odd stages use single inference:
\[
S_\text{single}(x; \mathcal{M}_i) = \bigl( \hat{y}, \Phi_i(x), \Xi_i(x) \bigr),
\]
where $\mathcal{M}_i \in \{\mathcal{M}_\text{base}, \mathcal{M}_\text{large}\}$ denotes model scale $i$. 
Even stages activate deliberation only on marginal confidence:
\[
S_\text{multi}(x; \mathcal{M}_i, N) = \text{Consensus}\bigl( \{\mathcal{M}_i(x; r_j)\}_{j=1}^N \bigr),
\]
where $r_j$ are distinct role prompts and Consensus aggregates via majority vote with confidence $\Phi_\text{multi}(x) = \mathbb{P}(\text{agreement})$.

This design scales test-time compute intra-tier before inter-tier escalation. 
Marginal cases trigger $S_\text{multi}$ to resolve ambiguities via emergent cooperation, delaying costly upgrades. 
The architecture naturally extends to additional model scales by repeating the single$\to$multi pattern.

\subsection{Confidence Estimation}
\label{sec:confidence}

CascadeDebate employs complementary confidence signals tailored to stage type, followed by Bayesian calibration. 
For single-model stages $S_\text{single}$, we extract $\Phi_k(x)$ via surrogate token probability~\citep{kadavath2022language}, capturing model self-assessment of answer quality. 
For multi-agent stages $S_\text{multi}$, confidence reflects ensemble agreement: $\Phi_k(x) = \phi_k^\text{agree}$ of $N$ agents $\mathcal{M}_i(x; r_j)$.
The majority vote prediction and the agreement confidence are calculated as:
\begin{equation}
    \hat{y}_k^\text{mv} = \arg\max_a \sum_{j=1}^N \mathds{1}[\hat{y}_k^{(j)} = a]
    \label{eq:majority}
\end{equation}
\begin{equation}
    \phi_k^\text{agree} = \frac{|\{j : \hat{y}_k^{(j)} = \hat{y}_k^\text{mv}\}|}{N}
    \label{eq:agreement}
\end{equation}
Multi-agent deliberation provides robust intra-tier signal where high agreement rate resolved ambiguity without escalation.

Both signals undergo Bayesian logistic regression calibration, fitted on 100 held-out samples per model scale $\mathcal{M}_i$. 
Calibration ensures $\Phi_k(x) \approx \mathbb{P}(\hat{y}_k = y^* | x)$ across operating ranges.

\subsection{Online Threshold Learning}
\label{sec:online_learning}

Thresholds $\boldsymbol{\tau} = \{\tau_k^d, \tau_k^a\}_k$ govern routing, i.e. to accept if $\Phi_k(x) > \tau_k^d$, abstain if $\Xi_k(x) > \tau_k^a$. 
Following the setup of~\cite{fanconi2025cascaded}, we parameterize it via sigmoid: $\tau_k = \sigma(\theta_k)$ with $\theta_k \in \mathbb{R}$. 
Soft gating enables gradient optimization: accept probability $\pi_k = \sigma(\gamma . (\Phi_k(x) - \tau_k))$ where $\gamma=5$ controls sharpness.
Stage $k$ stopping probability chains prior deferrals: $p_k = \pi_k \prod_{i<k} (1-\pi_i)$. 
The multi-objective loss balances error and cost:
\begin{align}
    \mathcal{L}(\boldsymbol{\theta}) &= (1 - \mathbb{E}[p_k \cdot \text{corr}_k]) + \lambda \cdot \mathbb{E}[p_k \cdot c_k] 
    \label{eq:online_loss}
\end{align}
where $\mathbb{E}[\cdot]$ aggregates over reached stages, $\text{corr}_k = \mathds{1}[\hat{y}_k = y^*]$, and $c_\text{expert}$ assumes perfection.
Parameters update via Adam on mini-batches from replay buffer $\mathcal{B}$, accumulating human feedback post-deployment.
More details in Appendix~\ref{app:online}.

\section{Experiments and Results}
\label{sec:experiments_results}

\begin{table*}[t]
\centering
\caption{\textbf{CascadeDebate vs.\ Baselines.} Accuracy (\%) of Llama-3.2-Instruct (1B/3B) and Qwen2.5-Instruct (1.5B/3B) across single-model ($S_\text{single}(\mathcal{M}_\text{base/large})$), multi-agent ($S_\text{multi}(\mathcal{M}_\text{base/large})$), standard cascade ($S_\text{single}(\mathcal{M}_\text{base}) \to S_\text{single}(\mathcal{M}_\text{large})\to S_\text{human}$), and CascadeDebate ($S_\text{single}(\mathcal{M}_\text{base}) \to S_\text{multi}(\mathcal{M}_\text{base}) \to S_\text{single}(\mathcal{M}_\text{large}) \to S_\text{multi}(\mathcal{M}_\text{large})\to S_\text{human}$) systems. Best in \textbf{bold}.}
\label{tab:main_results}
\resizebox{0.95\textwidth}{!}{%
\begin{tabular}{l  cccccc}
\toprule
\multirow{2}{*}{\textbf{Dataset}} & \multicolumn{4}{c}{\textbf{Baselines}} & \multicolumn{2}{c}{\textbf{Cascade}} \\
\cmidrule(lr){2-5} \cmidrule(lr){6-7}
 & \textbf{$S_\text{single}(\mathcal{M}_\text{base})$} & \textbf{$S_\text{multi}(\mathcal{M}_\text{base})$} & \textbf{$S_\text{single}(\mathcal{M}_\text{large})$} & \textbf{$S_\text{multi}(\mathcal{M}_\text{large})$} & \textbf{Standard} & \textbf{Ours} \\
\midrule
\toprule
&\multicolumn{6}{c}{\textbf{Llama-3.2} (Base: 1B, Large: 3B)} \\
\midrule
\textbf{ARC-Easy}      & 68.56 & 73.89 & 89.67 & 91.00 & 93.90 & \textbf{95.33$\pm$0.703} \\
\textbf{ARC-Challenge} & 50.67 & 54.78 & 78.33 & 81.67 & 84.30 & \textbf{92.89$\pm$0.857} \\
\textbf{MMLU}          & 44.22 & 47.56 & 62.44 & 64.78 & 67.70 & \textbf{82.67$\pm$1.262} \\
\textbf{MedQA}         & 34.22 & 35.11 & 60.89 & 64.00 & 68.20 & \textbf{86.44$\pm$1.141} \\
\textbf{MedMCQA}       & 36.11 & 41.78 & 52.67 & 55.33 & 64.70 & \textbf{76.33$\pm$1.417} \\
\toprule
&\multicolumn{6}{c}{\textbf{Qwen2.5} (Base: 1.5B, Large: 3B)} \\
\midrule
\textbf{ARC-Easy}      & 83.44 & 88.44 & 93.44 & 94.33 & \textbf{96.20} & 92.00$\pm$0.009 \\
\textbf{ARC-Challenge} & 71.89 & 74.67 & 82.78 & 84.44 & \textbf{89.80} & 85.78$\pm$0.012 \\
\textbf{MMLU}          & 58.78 & 58.44 & 65.00 & 67.11 & 73.20 & \textbf{78.00$\pm$0.014} \\
\textbf{MedQA}         & 34.44 & 36.89 & 40.89 & 41.44 & 52.70 & \textbf{75.22$\pm$0.014} \\
\textbf{MedMCQA}       & 37.33 & 42.00 & 46.44 & 46.67 & 58.70 & \textbf{69.89$\pm$0.015} \\
\bottomrule
\end{tabular}%
}
\end{table*}

\begin{figure*}[t]
    \centering
    \includegraphics[width=\textwidth]{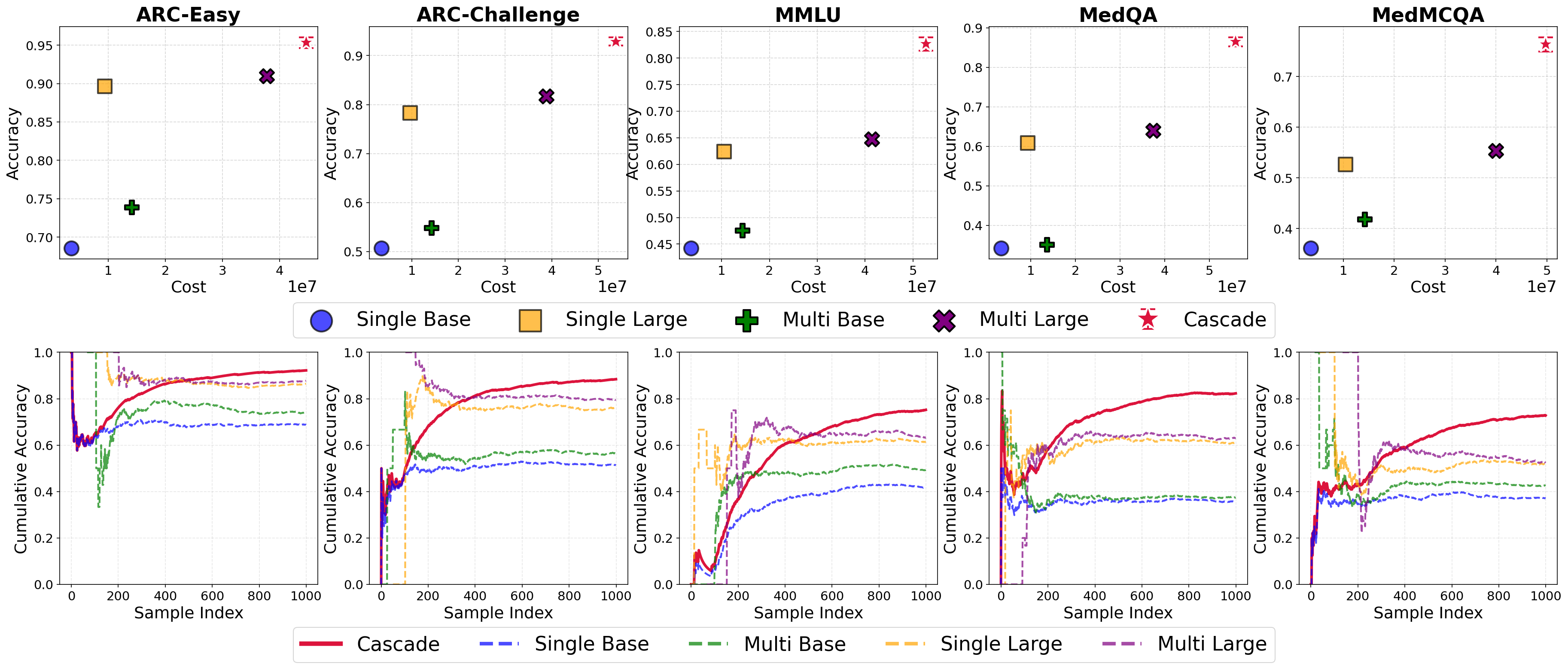}
    \caption{\textbf{CascadeDebate Cost-Accuracy Pareto Frontier.} 
    \textbf{Top:} Cost-accuracy curves across five benchmarks using Llama-3.2-Instruct model. 
    CascadeDebate (red $\star$) dominates the Pareto frontier, matching or exceeding $S_\text{multi}(\mathcal{M}_\text{large})$ accuracy at 20-35\% lower token cost. 
    \textbf{Bottom:} Online threshold adaptation. 
    CascadeDebate (solid red) rapidly surpasses fixed-threshold cascade (dashed red) and static baselines as the optimizer calibrates $\boldsymbol{\tau}_k$ to task difficulty over 1,000 samples.}

    \label{fig:main_results_plot}
\end{figure*}

The full cascade is
\[
\begin{aligned}
C &= S_\text{single}(\mathcal{M}_\text{base}) \to S_\text{multi}(\mathcal{M}_\text{base}) \\
&\quad\to S_\text{single}(\mathcal{M}_\text{large}) \to S_\text{multi}(\mathcal{M}_\text{large}) \\
&\quad\to S_\text{human}
\end{aligned}
\]
terminating early per thresholds $\boldsymbol{\tau} = \{\tau_k^d, \tau_k^a\}_k$.

\subsection{Setup details}
\label{sec:setup}
We evaluate CascadeDebate on five multiple-choice benchmarks, sampling 1,000 instances from each. 
These span science (ARC-Easy and ARC-Challenge~\cite{clark2018think}), general knowledge (MMLU~\cite{hendrycks2021measuring}), and medicine (MedQA~\cite{jin2021disease}; MedMCQA~\cite{pal2022medmcqa}). 
We use test splits for ARC and MMLU, validation splits for medical datasets, with statistics in Appendix~\ref{app:datasets}.
We employ instruction-tuned Llama-3.2~\cite{dubey2024llama} and Qwen2.5~\cite{qwen2.5}. 
Base models are Llama-3.2-1B-Instruct and Qwen2.5-1.5B-Instruct, while large models are Llama-3.2-3B-Instruct and Qwen2.5-3B-Instruct. 
All generation used $temp=0$ during inference with maximum length of 512 tokens.

Each dataset uses four specialized role prompts per multi-agent stage~(4 agents). 
For e.g., science tasks include \textit{Experimental Scientist} and \textit{Misconception Detector}; medical tasks include \textit{Clinical Reasoning} and \textit{Evidence-Based Medicine} (full prompts in Appendix~\ref{app:roles}).
All experiments run on a single NVIDIA A100 (80GB) GPU using PyTorch 2.4 and Transformers v4.44.2. 
Thresholds optimize via Adam (batch size 10, $\eta_\tau = 0.05$), initializing to $\tau_k^d = 0.6$. 
Hyperparameters are $\lambda=10^{-7}$ (cost-accuracy trade-off), $\rho=5.0$ (output-to-input price ratio), and $c_\text{expert}=10.0$.

\subsection{Results \& Observations}
\label{sec:results}

CascadeDebate answers two key questions across five benchmarks:
\begin{enumerate}
    \item Does embedding multi-agent deliberation at escalation boundaries outperform individual-stage baselines?
    \item How does the system allocate compute across capability tiers?
\end{enumerate}

We compare CascadeDebate against single-model, multi-agent, and standard cascade baselines across model scales (Table~\ref{tab:main_results}). 
A consistent trend is that multi-agent variants outperform their single-model counterparts, and larger models further improve accuracy within each dataset. 
Standard cascades that route from $S_\text{single}(\mathcal{M}_\text{base}) \to S_\text{single}(\mathcal{M}_\text{large}) \to S_{\text{human}}$ already surpass any individual stage, confirming the value of confidence-based deferral even without intra-tier deliberation.

\noindent\textbf{Superior accuracy at controlled cost.}
CascadeDebate achieves the best accuracy on all tasks with Llama-3.2, improving over the strongest baseline by 1.43-18.24 percentage points (pp) (e.g., MedQA: 86.44\% vs.\ 64.00\% for $S_\text{multi}(\mathcal{M}_\text{large})$ and 68.20\% for the standard cascade). 
For Qwen2.5, CascadeDebate matches or exceeds the standard cascade on three of five datasets and yields especially large gains on the medical benchmarks (e.g., MedQA: 75.22\% vs.\ 52.70\%). 
These patterns show that inserting deliberation at escalation boundaries systematically boosts accuracy beyond what can be obtained from scaling model size or naïve cascades alone.

\noindent\textbf{Effect of model scale and deliberation.}
Within each backbone, moving from $S_\text{single}(\mathcal{M}_\text{base})$ to $S_\text{single}(\mathcal{M}_\text{large})$ yields substantial gains (e.g., Llama-3.2 MMLU: 44.22\% $\to$ 62.44\%), and adding multi-agent deliberation on top of a fixed scale yields further improvements (62.44\% $\to$ 64.78\% for $S_\text{multi}(\mathcal{M}_\text{large})$). 
However, CascadeDebate consistently outperforms both single and multi-agent large models (e.g., Llama-3.2 MedMCQA: 55.33\% vs.\ 76.33\%), indicating that selective intra-tier cooperation is more effective than uniformly applying multi-agent methods at a single scale.

\noindent\textbf{Benefit over standard cascades.}
Standard cascades already close much of the gap between base and large models (e.g., Qwen2.5 ARC-Challenge: 71.89\% $\to$ 89.80\%), demonstrating that confidence-based routing is a strong baseline. 
CascadeDebate nonetheless delivers additional gains, most notably on harder and medical benchmarks where ambiguity is common. 
These improvements support our hypothesis that many escalations can be resolved by intra-tier debate rather than blindly deferring to larger or human solvers.

\noindent\textbf{Implications for compute allocation.}
Detailed cost-accuracy curves are shown in Fig.~\ref{fig:main_results_plot}. 
CascadeDebate systematically converts additional compute into disproportionate accuracy gains across domains.
On relatively easier ARC-Easy (Llama-3.2), CascadeDebate delivers +26.78pp (68.56\% $\to$ {95.33\%}) at only 12.79$\times$ Single Base cost, surpassing even $S_\text{multi}(\mathcal{M}_\text{large})$ (91.00\%) while selectively activating lightweight base deliberation.
For complex reasoning in ARC-Challenge, cost rises to 15.62$\times$ but yields massive +42.22pp (50.67\% $\to$ {92.89\%}), 1.62$\times$ the gain of standard cascade (76.78\%). 
The most compelling results appear on medical benchmarks: MedQA achieves +52.22pp (34.22\% $\to$ {86.44\%}) at 16.66$\times$ cost; MedMCQA gains +40.22pp (36.11\% $\to$ {76.33\%}) at 13.71$\times$. 

These trends confirm the core hypothesis: intra-tier multi-agent deliberation resolves good amount of escalation cases internally, concentrating expensive large-model and $S_\text{human}$ compute precisely where single models fail catastrophically. 
The system processes straightforward queries efficiently via $S_\text{single}(\mathcal{M}_\text{base})$ while reserving intensive reasoning for true edge cases, achieving cost-accuracy trade-offs.
Similar patterns hold for Qwen2.5 (see Appendix~\ref{app:qwen}, Fig. ~\ref{fig:app_qwen_results}).
\section{Discussion}
\label{sec:discussion}

CascadeDebate demonstrates that carefully orchestrated small models can surpass significantly larger standalone systems.
Here, we examine the mechanisms driving this performance, the role of adaptive thresholding, practical implications, and key limitations.

\noindent\textbf{Inference-time compute trumps parametric scaling.} 
Our 1B-parameter online cascade consistently outperforms 3B standalone models by 14.56-23.66\,pp across challenging benchmarks (ARC-Challenge: 92.89\% vs.\ 78.33\%; MedMCQA: 76.33\% vs.\ 52.67\%). 
This challenges conventional scaling laws by showing that inference-time compute, through selective multi-agent deliberation, can generate more effective reasoning than additional parameters alone. 
Selective multi-agent deliberation rectifies marginal uncertainties internally, enabling compact models to overcome parametric limitations through targeted test-time compute rather than uniform redundancy.

\noindent\textbf{Online adaptation is critical.} 
Fixed thresholds ($\tau_k=0.6$) perform reasonably but plateau between base and large model performance, as a single cutoff proves suboptimal across domains. 
Our online optimizer transforms this limitation, learning task-specific boundaries that balance frugality and safety by processing 44.40\% of ARC-Easy queries at the first stage while aggressively escalating only true edge cases in medical domains.
Refer to Appendix~\ref{app:analysis} (Table~\ref{tab:ablation_online}) for comparison of fixed to learned threshold.

\noindent\textbf{Elastic resource allocation.} 
The architecture exhibits confidence-driven elasticity. It incurs 12.79$\times$ Single Base cost on ARC-Easy versus 15.62$\times$ on ARC-Challenge, yielding accuracy gains of 26.78pp to 42.22pp respectively.
This adaptability delivers Pareto-superior cost-accuracy tradeoffs, concentrating expensive human and large-model compute precisely where single models fail catastrophically.
We report stage-wise distributions in Appendix~\ref{app:analysis} (Fig.~\ref{fig:app_dynamics}).

Human expert cost modeling assumes perfect accuracy, which may overestimate final-stage performance in practice.
Future work should explore automated role discovery, test-time distillation of consensus reasoning, and deployment on long-context production workloads.

\section{Conclusion}
\label{sec:conclusion}


CascadeDebate embeds multi-agent deliberation at cascade escalation boundaries, alternating single-model inference with selective agent ensembles across scales, with human experts as final fallback.
Confidence-based routers activate consensus only for marginal cases, resolving ambiguities internally before costly escalation.
Our online threshold optimizer transforms static cascades into elastic reasoners, adapting to query distributions while balancing accuracy against compute costs. 
This inference-time scaling outperforms traditional parametric approaches across diverse domains.
By concentrating expensive compute where single models fail most catastrophically, CascadeDebate delivers superior cost-accuracy tradeoffs with human-like cognitive elasticity-lightweight for routine queries and intensive deliberation for edge cases.
Future work will explore parallel execution, automated role discovery, and extension to open-ended generation.

\newpage
\section*{Limitations}
\label{sec:limitations}

We acknowledge three limitations. First, latency arises from the sequential nature that introduces cumulative delays that exceed single-pass models. Second, error propagation occurs when miscalibrated base models prematurely accept incorrect answers and prevent the necessary escalation. Third, agent homogeneity limits diversity compared to heterogeneous ensembles as we rely solely on role-prompting within a single model family.



\bibliography{custom}

@article{clark2018think,
  title={Think you have solved question answering? try arc, the ai2 reasoning challenge},
  author={Clark, Peter and Cowhey, Isaac and Etzioni, Oren and Khot, Tushar and Sabharwal, Ashish and Schoenick, Carissa and Tafjord, Oyvind},
  journal={arXiv preprint arXiv:1803.05457},
  year={2018}
}

@inproceedings{
hendrycks2021measuring,
title={Measuring Massive Multitask Language Understanding},
author={Dan Hendrycks and Collin Burns and Steven Basart and Andy Zou and Mantas Mazeika and Dawn Song and Jacob Steinhardt},
booktitle={International Conference on Learning Representations},
year={2021},
url={https://openreview.net/forum?id=d7KBjmI3GmQ}
}

@article{jin2021disease,
  title={What disease does this patient have? a large-scale open domain question answering dataset from medical exams},
  author={Jin, Di and Pan, Eileen and Oufattole, Nassim and Weng, Wei-Hung and Fang, Hanyi and Szolovits, Peter},
  journal={Applied Sciences},
  volume={11},
  number={14},
  pages={6421},
  year={2021},
  publisher={MDPI}
}

@inproceedings{pal2022medmcqa,
  title={Medmcqa: A large-scale multi-subject multi-choice dataset for medical domain question answering},
  author={Pal, Ankit and Umapathi, Logesh Kumar and Sankarasubbu, Malaikannan},
  booktitle={Conference on health, inference, and learning},
  pages={248--260},
  year={2022},
  organization={PMLR}
}

@article{singhal2023large,
  title={Large language models encode clinical knowledge},
  author={Singhal, Karan and Azizi, Shekoofeh and Tu, Tao and Mahdavi, S Sara and Wei, Jason and Chung, Hyung Won and Scales, Nathan and Tanwani, Ajay and Cole-Lewis, Heather and Pfohl, Stephen and others},
  journal={Nature},
  volume={620},
  number={7972},
  pages={172--180},
  year={2023},
  publisher={Nature Publishing Group}
}

@article{
chen2024frugalgpt,
title={Frugal{GPT}: How to Use Large Language Models While Reducing Cost and Improving Performance},
author={Lingjiao Chen and Matei Zaharia and James Zou},
journal={Transactions on Machine Learning Research},
issn={2835-8856},
year={2024},
url={https://openreview.net/forum?id=cSimKw5p6R},
note={Featured Certification}
}

@inproceedings{
yue2024large,
title={Large Language Model Cascades with Mixture of Thought Representations for Cost-Efficient Reasoning},
author={Murong Yue and Jie Zhao and Min Zhang and Liang Du and Ziyu Yao},
booktitle={The Twelfth International Conference on Learning Representations},
year={2024},
url={https://openreview.net/forum?id=6okaSfANzh}
}

@article{ji2023survey,
  title={Survey of hallucination in natural language generation},
  author={Ji, Ziwei and Lee, Nayeon and Frieske, Rita and Yu, Tiezheng and Su, Dan and Xu, Yan and Ishii, Etsuko and Bang, Ye Jin and Madotto, Andrea and Fung, Pascale},
  journal={ACM computing surveys},
  volume={55},
  number={12},
  pages={1--38},
  year={2023},
  publisher={ACM New York, NY}
}

@article{wei2022chain,
  title={Chain-of-thought prompting elicits reasoning in large language models},
  author={Wei, Jason and Wang, Xuezhi and Schuurmans, Dale and Bosma, Maarten and Xia, Fei and Chi, Ed and Le, Quoc V and Zhou, Denny and others},
  journal={Advances in neural information processing systems},
  volume={35},
  pages={24824--24837},
  year={2022}
}

@article{aggarwal2024automix,
  title={AutoMix: Automatically mixing language models},
  author={Aggarwal, Pranjal and Madaan, Aman and Anand, Ankit and Potharaju, Srividya Pranavi and Mishra, Swaroop and Zhou, Pei and Gupta, Aditya and Rajagopal, Dheeraj and Kappaganthu, Karthik and Yang, Yiming and others},
  journal={Advances in Neural Information Processing Systems},
  volume={37},
  pages={131000--131034},
  year={2024}
}

@inproceedings{
ong2025routellm,
title={Route{LLM}: Learning to Route {LLM}s from Preference Data},
author={Isaac Ong and Amjad Almahairi and Vincent Wu and Wei-Lin Chiang and Tianhao Wu and Joseph E. Gonzalez and M Waleed Kadous and Ion Stoica},
booktitle={The Thirteenth International Conference on Learning Representations},
year={2025},
url={https://openreview.net/forum?id=8sSqNntaMr}
}

@inproceedings{
du2024improving,
title={Improving Factuality and Reasoning in Language Models through Multiagent Debate},
author={Yilun Du and Shuang Li and Antonio Torralba and Joshua B. Tenenbaum and Igor Mordatch},
booktitle={Forty-first International Conference on Machine Learning},
year={2024},
url={https://openreview.net/forum?id=zj7YuTE4t8}
}

@inproceedings{
ding2024hybrid,
title={Hybrid {LLM}: Cost-Efficient and Quality-Aware Query Routing},
author={Dujian Ding and Ankur Mallick and Chi Wang and Robert Sim and Subhabrata Mukherjee and Victor R{\"u}hle and Laks V. S. Lakshmanan and Ahmed Hassan Awadallah},
booktitle={The Twelfth International Conference on Learning Representations},
year={2024},
url={https://openreview.net/forum?id=02f3mUtqnM}
}

@article{
li2024more,
title={More Agents Is All You Need},
author={junyou li and Qin Zhang and Yangbin Yu and QIANG FU and Deheng Ye},
journal={Transactions on Machine Learning Research},
issn={2835-8856},
year={2024},
url={https://openreview.net/forum?id=bgzUSZ8aeg},
note={}
}

@article{li2023camel,
  title={Camel: Communicative agents for" mind" exploration of large language model society},
  author={Li, Guohao and Hammoud, Hasan and Itani, Hani and Khizbullin, Dmitrii and Ghanem, Bernard},
  journal={Advances in Neural Information Processing Systems},
  volume={36},
  pages={51991--52008},
  year={2023}
}

@inproceedings{
wang2025mixtureofagents,
title={Mixture-of-Agents Enhances Large Language Model Capabilities},
author={Junlin Wang and Jue WANG and Ben Athiwaratkun and Ce Zhang and James Zou},
booktitle={The Thirteenth International Conference on Learning Representations},
year={2025},
url={https://openreview.net/forum?id=h0ZfDIrj7T}
}

@article{dekoninck2024unified,
  title={A unified approach to routing and cascading for llms},
  author={Dekoninck, Jasper and Baader, Maximilian and Vechev, Martin},
  journal={arXiv preprint arXiv:2410.10347},
  year={2024}
}

@inproceedings{shen2025sater,
  title={SATER: A Self-Aware and Token-Efficient Approach to Routing and Cascading},
  author={Shen, Yuanzhe and Liu, Yide and Huang, Zisu and Yin, Ruicheng and Zheng, Xiaoqing and Huang, Xuan-Jing},
  booktitle={Proceedings of the 2025 Conference on Empirical Methods in Natural Language Processing},
  pages={10526--10540},
  year={2025}
}

@article{kadavath2022language,
  title={Language models (mostly) know what they know},
  author={Kadavath, Saurav and Conerly, Tom and Askell, Amanda and Henighan, Tom and Drain, Dawn and Perez, Ethan and Schiefer, Nicholas and Hatfield-Dodds, Zac and DasSarma, Nova and Tran-Johnson, Eli and others},
  journal={arXiv preprint arXiv:2207.05221},
  year={2022}
}

@inproceedings{
xiong2024can,
title={Can {LLM}s Express Their Uncertainty? An Empirical Evaluation of Confidence Elicitation in {LLM}s},
author={Miao Xiong and Zhiyuan Hu and Xinyang Lu and YIFEI LI and Jie Fu and Junxian He and Bryan Hooi},
booktitle={The Twelfth International Conference on Learning Representations},
year={2024},
url={https://openreview.net/forum?id=gjeQKFxFpZ}
}

@article{zhang2024chain,
  title={Chain of agents: Large language models collaborating on long-context tasks},
  author={Zhang, Yusen and Sun, Ruoxi and Chen, Yanfei and Pfister, Tomas and Zhang, Rui and Arik, Sercan},
  journal={Advances in Neural Information Processing Systems},
  volume={37},
  pages={132208--132237},
  year={2024}
}

@inproceedings{SunZW025_cortex,
  author={Yiliu Sun and Zicheng Zhao and Sheng Wan and Chen Gong},
  title={CortexDebate: Debating Sparsely and Equally for Multi-Agent Debate},
  year={2025},
  cdate={1735689600000},
  pages={9503-9523},
  url={https://aclanthology.org/2025.findings-acl.495/},
  booktitle={ACL (Findings)}
}

@article{chen2025multi,
  title={Multi-agent evolve: Llm self-improve through co-evolution},
  author={Chen, Yixing and Wang, Yiding and Zhu, Siqi and Yu, Haofei and Feng, Tao and Zhang, Muhan and Patwary, Mostofa and You, Jiaxuan},
  journal={arXiv preprint arXiv:2510.23595},
  year={2025}
}

@article{dubey2024llama,
  title={The llama 3 herd of models},
  author={Dubey, Abhimanyu and Jauhri, Abhinav and Pandey, Abhinav and Kadian, Abhishek and Al-Dahle, Ahmad and Letman, Aiesha and Mathur, Akhil and Schelten, Alan and Yang, Amy and Fan, Angela and others},
  journal={arXiv e-prints},
  pages={arXiv--2407},
  year={2024}
}

@article{qwen2.5,
    title   = {Qwen2.5 Technical Report}, 
    author  = {An Yang and Baosong Yang and Beichen Zhang and Binyuan Hui and Bo Zheng and Bowen Yu and Chengyuan Li and Dayiheng Liu and Fei Huang and Haoran Wei and Huan Lin and Jian Yang and Jianhong Tu and Jianwei Zhang and Jianxin Yang and Jiaxi Yang and Jingren Zhou and Junyang Lin and Kai Dang and Keming Lu and Keqin Bao and Kexin Yang and Le Yu and Mei Li and Mingfeng Xue and Pei Zhang and Qin Zhu and Rui Men and Runji Lin and Tianhao Li and Tingyu Xia and Xingzhang Ren and Xuancheng Ren and Yang Fan and Yang Su and Yichang Zhang and Yu Wan and Yuqiong Liu and Zeyu Cui and Zhenru Zhang and Zihan Qiu},
    journal = {arXiv preprint arXiv:2412.15115},
    year    = {2024}
}

@inproceedings{lee2025gemmas,
  title={Gemmas: Graph-based evaluation metrics for multi agent systems},
  author={Lee, Jisoo and Chang, Raeyoung and Kwon, Dongwook and Singh, Harmanpreet and Verma, Nikhil},
  booktitle={Proceedings of the 2025 Conference on Empirical Methods in Natural Language Processing: Industry Track},
  pages={1522--1532},
  year={2025}
}

@article{fanconi2025cascaded,
  title={Cascaded Language Models for Cost-effective Human-AI Decision-Making},
  author={Fanconi, Claudio and van der Schaar, Mihaela},
  journal={arXiv preprint arXiv:2506.11887},
  year={2025}
}
\clearpage
\appendix
\raggedbottom 

\section{Online Learning Derivation and Algorithm}
\label{app:online}
The confidence $\Phi_k(x)$ and uncertainty $\Xi_k(x)$ metrics used in \textit{CascadeDebate} are designed to provide complementary signals for intra-tier and inter-tier escalation. For single-model stages, we utilize the \textit{Surrogate Token Probability} (STP), whereas for multi-agent stages, $\Phi_k$ is defined as the \textit{Agreement Rate} among $N$ agents. Both are naturally bounded as $\Phi_k(x) \in [0, 1]$, representing a probabilistic estimate of correctness. In contrast, the uncertainty metric $\Xi_k(x) \in [0, \infty)$ quantifies the ``unknown unknowns'' or epistemic uncertainty, typically computed using the logit entropy of the base model or the variance across agent outputs. While clear queries yield values near 0, highly ambiguous inputs result in high entropy that theoretically approaches infinity, triggering an escalation to human experts. To ensure these raw signals are reliable, we apply \textit{Bayesian Logistic Regression} fitted on a held-out calibration set. This process maps raw scores to empirical accuracy $P(\hat{y}_k = y^* | x)$, correcting for the typical over-confidence observed in large language models.

Since the decision to accept or defer at each stage is discrete (based on the indicator function $\mathbb{I}[\phi_k \ge \tau_k]$), the standard gradient descent cannot be applied directly. To enable end-to-end differentiability, we employ a soft relaxation technique during the backward pass.

We parameterize each threshold $\tau_k$ as a scalar $\theta_k \in \mathbb{R}$ such that $\tau_k = \sigma(\theta_k)$, ensuring that the threshold remains within $(0,1)$.
During optimization, we approximate the ``pass probability'' (probability of deferring to the next stage) using a shifted sigmoid function:
\begin{equation}
    \pi_k \approx \sigma\left(\kappa \cdot (\tau_k - \phi_k)\right)
\end{equation}
where $\kappa$ is a temperature scaling factor (set to 5.0) that controls the sharpness of the decision boundary.
Using this soft probability, the expected cumulative cost and expected error become differentiable functions of $\theta_k$.

The Algorithm~\ref{alg:cascade} details the inference and update loop. We maintain a replay buffer $\mathcal{B}$ to store recent query tuples $(\phi_{1:4}, y_{\text{true}}, c_{1:4})$. In update steps, we sample a mini-batch from $\mathcal{B}$ and perform a gradient update on $\{\theta_k\}$ to minimize compound loss $\mathcal{L}$ using the Adam optimizer.

\begin{algorithm}[h]
\caption{Five-Stage Cascade with Online Threshold Learning}
\label{alg:cascade}
\begin{algorithmic}[1]
\REQUIRE Query $x$, thresholds $\{\tau_k\}_{k=1}^{4}$, models $\mathcal{M}_{\text{base}}$, $\mathcal{M}_{\text{large}}$, role prompts $\{r_j\}_{j=1}^{N}$
\ENSURE Answer $\hat{y}$, cost $C$

\STATE $C \leftarrow 0$
\FOR{$k = 1$ \TO ~$4$}
\STATE Generate $\hat{y}_k$ using stage $S_k$ configuration
\STATE Compute confidence $\phi_k$
\STATE $C \leftarrow C + c_k$
\IF{$\phi_k \ge \tau_k$}
\STATE \textbf{return} $\hat{y}_k$, $C$ \COMMENT{Accept at stage $k$}
\ENDIF
\ENDFOR
\STATE \textbf{return} $y_{\text{expert}}$, $C + c_{\text{expert}}$ \COMMENT{Defer to human}
\vspace{0.3em}
\STATE \textit{// Online update (if enabled):}
\STATE Append $(\phi_{1:4},\, \mathds{1}[\hat{y}_k{=}y],\, c_{1:4})$ to $\mathcal{B}$
\STATE Update $\{\theta_k\}$ on mini-batch from $\mathcal{B}$
\end{algorithmic}
\end{algorithm}

\section{Dataset Statistics}
\label{app:datasets}

Table~\ref{tab:datasets} provides a detailed summary of the five benchmarks used in our evaluation. To ensure a consistent computational budget across all domains, we randomly sampled $N{=}1,000$ instances from each dataset.

For ARC and MMLU, we utilized the official test splits. However, for medical datasets (MedQA and MedMCQA), the test set labels are often withheld for leaderboard competitions; therefore, we performed evaluations on the validation splits.

\paragraph{Benchmarks Descriptions.}
\begin{itemize}
    \item ARC-Easy \& Challenge \citep{clark2018think}: Sourced from grade-school science exams. Easy questions are solvable via retrieval, whereas Challenge questions contain adversarial choices requiring multi-hop reasoning.
    \item MMLU \citep{hendrycks2021measuring}: A massive multitask benchmark covering 57 subjects (e.g., math, history, law) ranging from elementary to professional levels.
    \item MedQA \citep{jin2021disease}: Derived from the US Medical Licensing Examination (USMLE), which requires extensive professional knowledge and clinical reasoning.
    \item MedMCQA \citep{pal2022medmcqa}: Collected from Indian medical entrance exams (AIIMS, NEET), featuring high-difficulty multiple-choice questions.
\end{itemize}

\begin{table}[t]
    \centering
    \small
    \resizebox{\columnwidth}{!}{
    \begin{tabular}{llccc}
        \toprule
        Dataset & Domain & Split & Choices & Samples \\
        \midrule
        ARC-Easy & Science & Test & 3--5 & 1,000 \\
        ARC-Challenge & Science & Test & 3--5 & 1,000 \\
        MMLU & General & Test & 4 & 1,000 \\
        MedQA & Medical & Validation & 5 & 1,000 \\
        MedMCQA & Medical & Validation & 4 & 1,000 \\
        \bottomrule
    \end{tabular}
    }
    \caption{Summary of datasets. We use 1,000 randomly sampled instances from each dataset for consistent evaluation.}
    \label{tab:datasets}
\end{table}

\section{Role Prompts}
\label{app:roles}

Each stage of multiple agents ($S_2, S_4$) utilizes four domain-specific role triggers to induce diverse perspectives of reasoning. Table~\ref{tab:roles} lists the specific roles assigned for each data set. 
At inference time, each agent is instantiated with a unique role system prompt to generate an independent response, after which their outputs are aggregated via majority voting.

\begin{table}[t]
\centering
\small
\begin{tabular}{lp{5.5cm}}
\toprule
Dataset & Agent Roles \\
\midrule
ARC & Causal Chain Specialist, Misconception Detector, Experimental Scientist, Reviewer \\
\midrule
MedQA & Clinical Reasoning Expert, Evidence-Based Medicine Specialist, Option Eliminator, Safety Reviewer \\
\midrule
MedMCQA & High-Yield Facts Expert, Clinical Pattern Recognizer, Strategic Eliminator, Exam Strategist \\
\midrule
MMLU & First-Principles Thinker, Context Analyst, Consistency Checker, Textbook Expert \\
\bottomrule
\end{tabular}
\caption{Domain-specific agent roles used in the multi-agent stages.}
\label{tab:roles}
\end{table}

\section{Results on Qwen2.5}
\label{app:qwen}

\begin{figure*}[t!]
    \centering
    \includegraphics[width=0.95\textwidth]{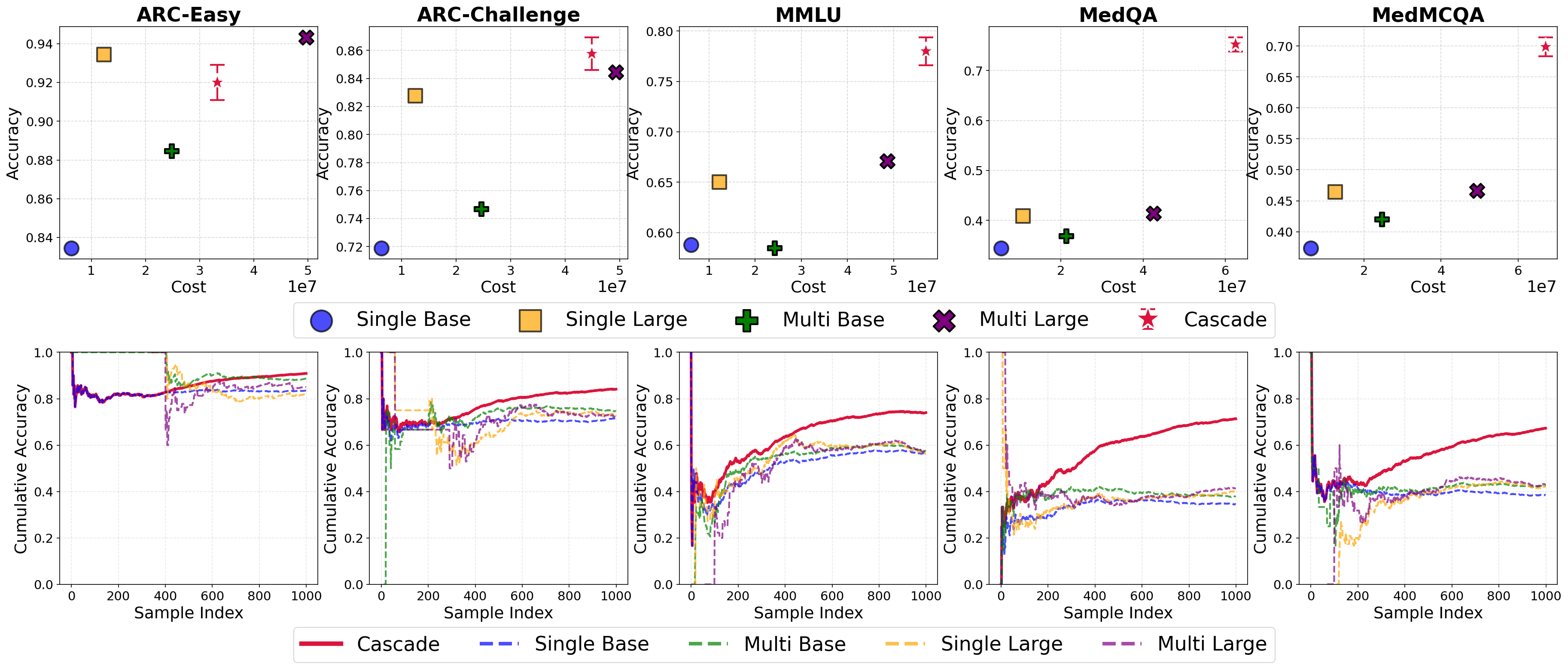}
    \caption{Generalizability Analysis using Qwen2.5 Models. 
    The performance trends are consistent with the Llama-3.2 results.
    Top Row: The \textit{Cascade} (red star $\star$) consistently occupies the optimal position on the Pareto frontier across all benchmarks.
    Bottom Row: The online threshold learning curve demonstrates rapid adaptation and stability, confirming that our proposed cost-aware optimization is robust regardless of the underlying backbone model.}
    \label{fig:app_qwen_results}
\end{figure*}

To verify the generalizability of our framework in different model architectures, we extended our experiments to the Qwen2.5 family \citep{qwen2.5}. Specifically, we designated \texttt{Qwen2.5-1.5B-Instruct} as the Base Model (Stages $S_1, S_2$) and \texttt{Qwen2.5-3B-Instruct} as the Large Model (Stages $S_3, S_4$).

Figure~\ref{fig:app_qwen_results} (Top Row) presents the cost-accuracy trade-off on five benchmarks. Consistent with the main results using Llama-3.2, the Qwen-based Cascade system (marked with a red star $\star$) successfully identifies the optimal operating point.
\begin{itemize}
    \item \textbf{Superiority over Static Baselines:} The cascade consistently outperforms the \textit{Single Base} and \textit{multi-agent-base} models in accuracy by a significant margin. Crucially, it achieves performance comparable to or exceeding the computationally expensive \textit{Multi-Agent Large} baseline while consuming significantly fewer tokens.
    \item \textbf{Robustness across Domains:} Whether in scientific reasoning (ARC) or medical knowledge (MedQA), the framework maintains a high position on the Pareto frontier.
\end{itemize}

The learning curves (Bottom Row) confirm the stability of our optimization algorithm in the Qwen architecture. The cumulative accuracy improves rapidly in the early phase (first 100--200 samples) as the thresholds calibrate to the data set's difficulty. This indicates that our proposed loss function and STP method are robust transferable components that function reliably across different LLM families.

\section{Additional Experimental Analysis}
\label{app:analysis}

In this section, we provide a deeper analysis of the cascade's internal dynamics and conduct an ablation study of the thresholding strategy. All analyzes presented here are based on the Llama-3.2 family (1B and 3B) as backbone models.

\subsection{Comparison with Fixed Thresholds}
\begin{figure*}[t!]
    \centering
    \includegraphics[width=0.95\textwidth]{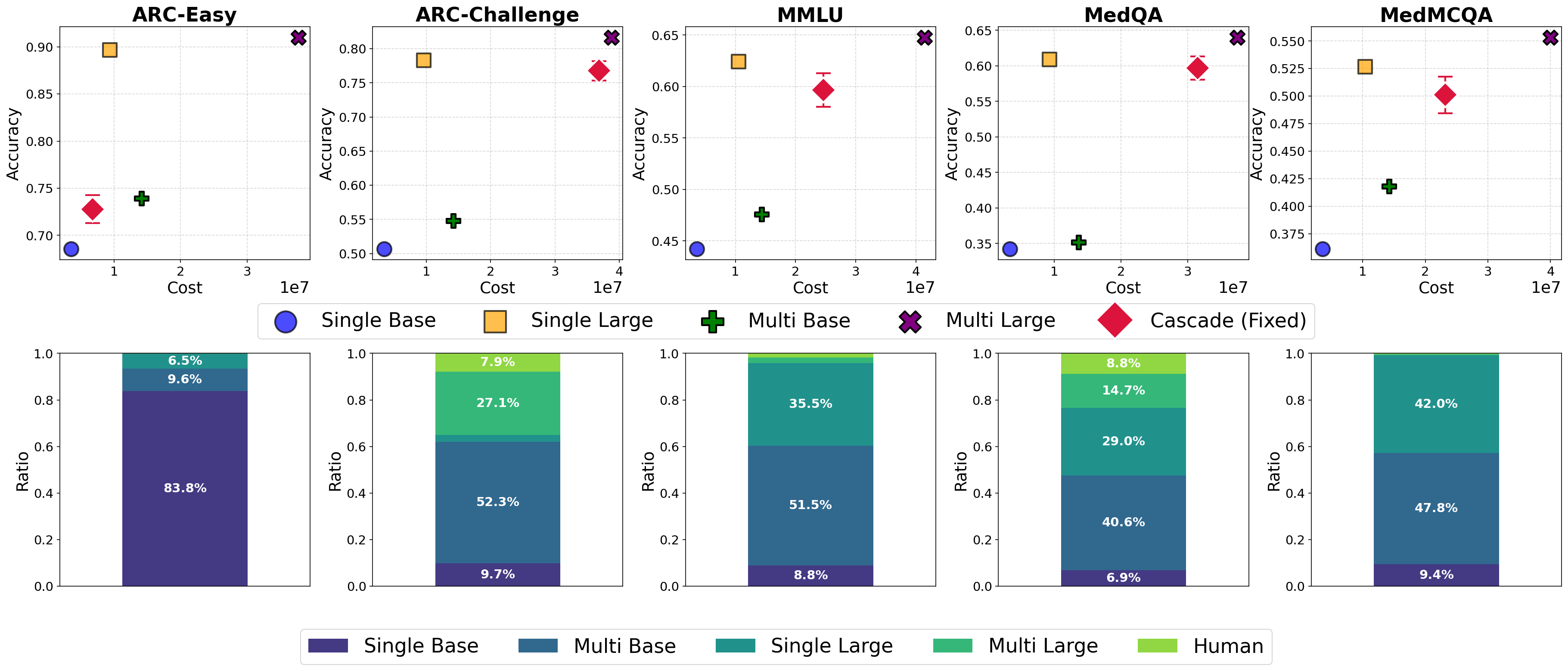}
    \caption{Performance Analysis of the Fixed Threshold Strategy on Llama-3.2 ($\tau=0.6$). 
    Top Row: Cost-Accuracy trade-off. The \textit{Fixed Cascade} (red diamond $\diamondsuit$) generally improves upon the Single Base model but often falls short of the optimal Pareto frontier compared to the Online strategy.
    Bottom Row: Stage distribution ratios. Without dynamic adjustment, the system relies on static confidence scores, resulting in a rigid allocation of resources.}
    \label{fig:app_fixed_analysis}
\end{figure*}

To validate the need for online optimization, we analyze the performance of a static cascade where all thresholds are fixed at $\tau=0.6$ (Figure~\ref{fig:app_fixed_analysis}).

Unlike the online approach, the \textit{Fixed Cascade} fails to achieve the optimal Pareto frontier. As shown in the stage distribution ratios (Bottom Row), a static threshold results in a rigid allocation policy—often over-spending on simple queries or under-utilizing capable models on hard queries.

To quantify the performance gap, Table~\ref{tab:ablation_online} compares the precision of the Fixed Threshold strategy ($\tau=0.6$) with our Online Learning approach. The results show that the online strategy produces substantial accuracy improvements ranging from +16.1\% to +26.7\%.

\begin{table}[h]
\centering
\caption{Impact of Online Threshold Optimization. Comparison of accuracy between the Fixed Threshold strategy and the Online Learning strategy.}
\label{tab:ablation_online}
\resizebox{\columnwidth}{!}{
\begin{tabular}{lccc}
\toprule
Dataset & Fixed Threshold & Online Learning & Improvement \\
\midrule
ARC-Easy      & 72.8\% & 95.3\% & +22.5\% \\
ARC-Challenge & 76.8\% & 92.9\% & +16.1\% \\
MMLU          & 59.7\% & 82.7\% & +23.0\% \\
MedQA         & 59.7\% & 86.4\% & +26.7\% \\
MedMCQA       & 50.1\% & 76.3\% & +26.2\% \\
\bottomrule
\end{tabular}%
}
\end{table}

\subsection{Cascade Dynamics and Elasticity}
\begin{figure*}[t]
    \centering
    \begin{subfigure}[t]{0.49\textwidth}
        \centering
        \includegraphics[width=\linewidth]{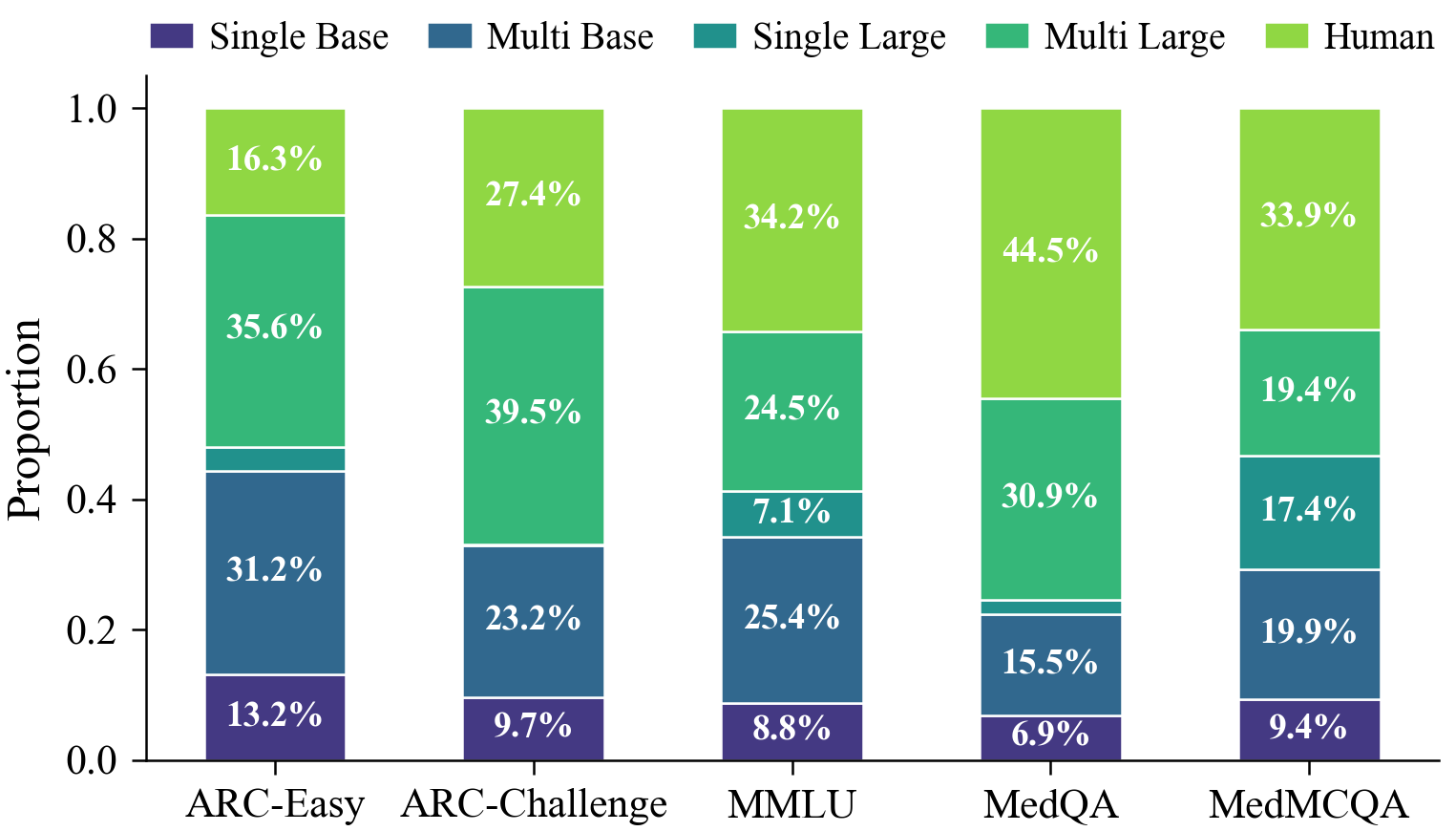}
        \caption*{(a) Llama-3.2}
    \end{subfigure}\hfill
    \begin{subfigure}[t]{0.49\textwidth}
        \centering
        \includegraphics[width=\linewidth]{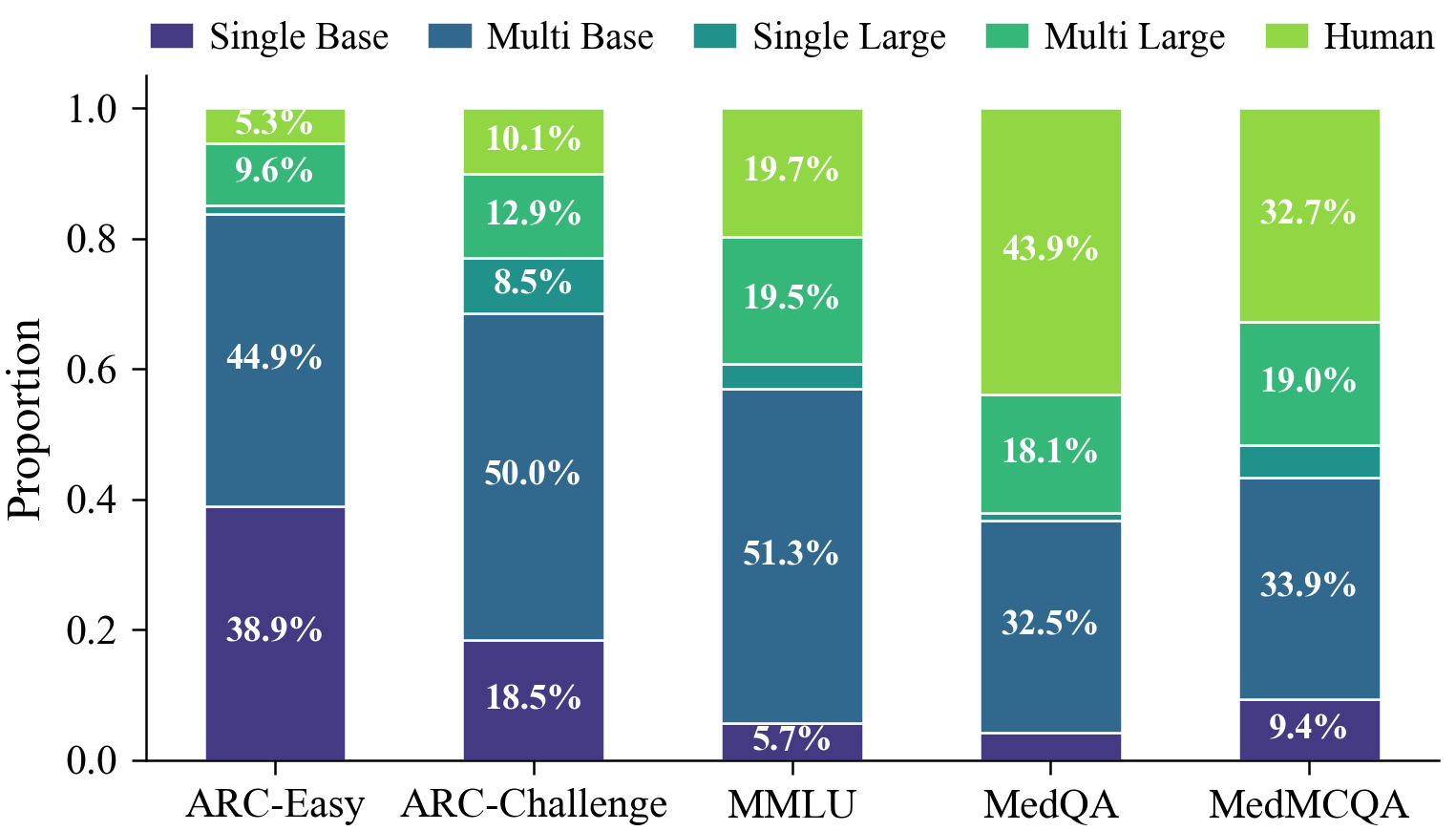}
        \caption*{(b) Qwen2.5}
    \end{subfigure}

    \caption{Stage-wise Sample Distribution with online learned thresholds for (a) Llama-3.2 and (b) Qwen2.5. Bars show the proportion of queries that terminate at each stage under confidence-based routing. Easier benchmarks are more likely to meet the learned acceptance thresholds in earlier stages and terminate without escalation. Harder benchmarks more frequently fall below early-stage thresholds, triggering deferral to later stages including larger models and the human expert.}
    \label{fig:app_dynamics}
\end{figure*}

Figure~\ref{fig:app_dynamics} shows the stage-wise distribution of query termination across benchmarks under online learned thresholds.
\begin{itemize}
    \item \textbf{Elastic Resource Allocation:} The cascade exhibits confidence-driven elasticity. On simpler benchmarks such as ARC-Easy, a larger fraction of queries terminate in earlier stages because the base model confidence more often exceeds the learned acceptance thresholds. On harder benchmarks such as MedQA, early-stage confidence more frequently falls below threshold, leading to increased deferral to later stages, including the multi-stages and the human expert stage.
\end{itemize}

\section{Human Involvement and Cascades}
Human involvement in AI systems broadly refers to settings where people participate in an otherwise automated decision process, for example, by reviewing uncertain outputs, correcting errors, or serving as final adjudicators for unresolved or high-risk cases. A cascade is a staged architecture that routes each input through solvers with different cost-capability profiles, typically starting with cheaper models and escalating only when confidence is low or the current stage abstains \citep{aggarwal2024automix, ong2025routellm}. The general goal is to match computation and oversight with query difficulty, so routine cases are handled efficiently, while ambiguous cases receive stronger models or human review \citep{fanconi2025cascaded}.

\end{document}